\documentclass{article}

\usepackage{PRIMEarxiv}

\usepackage[utf8]{inputenc} 
\usepackage[T1]{fontenc}    
\usepackage{hyperref}       
\usepackage{url}            
\usepackage{booktabs}       
\usepackage{amsfonts}       
\usepackage{nicefrac}       
\usepackage{microtype}      
\usepackage{lipsum}
\usepackage{fancyhdr}       
\usepackage{graphicx}       
\graphicspath{{media/}}     
\usepackage{multirow}
\usepackage{paralist}
\usepackage{enumitem}
\usepackage{amsmath}
\usepackage{float}

\title{GeneQuery: A General QA-based Framework for Spatial Gene Expression Predictions from Histology Images
}

\author{Ying Xiong\\
MBZUAI\\
\and
Linjing Liu \\
CUHK\\
\and
Yufei Cui \\
McGill University\\
\and
Shangyu WU \\
CityU\\
\and
Xue Liu \\
MBZUAI\\
\and
Antoni B. Chan \\
CityU\\
\and
Chun Jason Xue \\
MBZUAI\\
}

\begin{document}

\maketitle
\begin{abstract}
Gene expression profiling provides profound insights into molecular mechanisms, but its time-consuming and costly nature often presents significant challenges. 
In contrast, whole-slide hematoxylin and eosin (H\&E) stained histological images are readily accessible and allow for detailed examinations of tissue structure and composition at the microscopic level. 
Recent advancements have utilized these histological images to predict spatially resolved gene expression profiles. 
However, state-of-the-art works treat gene expression prediction as a multi-output regression problem, where each gene is learned independently with its own weights, failing to capture the shared dependencies and co-expression patterns between genes.
Besides, existing works can only predict gene expression values for genes seen during training, limiting their ability to generalize to new, unseen genes.

To address the above limitations, this paper presents GeneQuery, which aims to solve this gene expression prediction task in a question-answering (QA) manner for better generality and flexibility.
Specifically, GeneQuery takes gene-related texts as queries and whole-slide images as contexts and then predicts the queried gene expression values.
With such a transformation, GeneQuery can implicitly estimate the gene distribution by introducing the gene random variable.
Besides, the proposed GeneQuery consists of two architecture implementations, i.e., spot-aware GeneQuery for capturing patterns between images and gene-aware GeneQuery for capturing patterns between genes.
Comprehensive experiments on spatial transcriptomics datasets show that the proposed  GeneQuery outperforms existing state-of-the-art methods on known and unseen genes.
More results also demonstrate that GeneQuery can potentially analyze the tissue structure.
\end{abstract}


\section{Introduction}
\label{sec:intro}

\begin{figure}[t]
\centering
\includegraphics[width=0.6\linewidth]{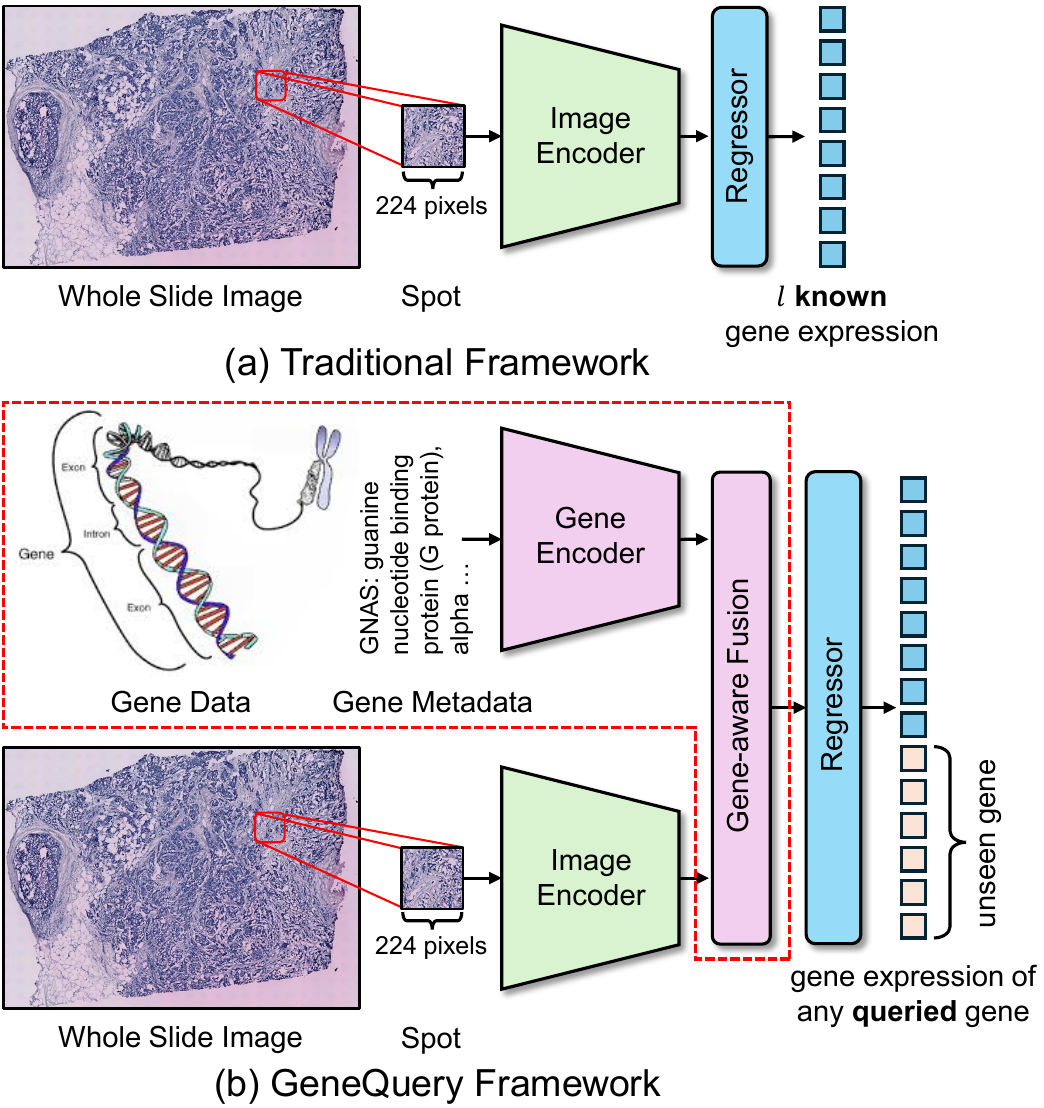}
\caption{Comparisons between GeneQuery and traditional framework. (a) traditional frameworks directly learn each gene's expression value as a multi-output regression task; (b) the proposed GeneQuery queries the metadata of given genes, including known and unseen genes, over the whole slide image.}
\label{fig:overview}
\end{figure}

Gene expression profiling provides a comprehensive view of the genetic activity within cells or organisms, which can be crucial for understanding various biological processes, disease mechanisms, and treatment responses. 
However, traditional gene expression profiling methods, such as bulk RNA sequencing, may not capture the heterogeneity within a sample, while single-cell methods capture heterogeneity without spatial context. 
Recently, spatial transcriptomics (ST) technologies have revolutionized our understanding of cellular and tissue-level biology by allowing for transcriptome-wide gene expression analysis while maintaining spatial context. 
Traditional ST methods, such as Visium~\cite{visuim}, MERFISH~\cite{merfish}, seqFISH+~\cite{seqfish+}, STARmap~\cite{STARmap}, smFISH~\cite{smFish}, and Targeted ExSeq~\cite{ExSeq}, have provided significant insights. 
However, these methods have inherent limitations, such as high cost, labor intensity, and platform-specific restrictions. 
On the other hand, histological images are relatively easy and inexpensive to obtain and offer a promising alternative for predicting transcript expression, potentially overcoming many obstacles current ST techniques face. 
These images provide rich, contextual information about tissue architecture and cellular morphology, which can be leveraged to infer spatial gene expression patterns. 

To generate large-scale ST data and assist in analyzing the molecular characteristics of tissues, recent works predicted spatial gene expression from histology images and provide the probability to generate gene expression from histology images, such as STNet~\cite{stnet}, HistoGene~\cite{histogene}, and BLEEP~\cite{bleep}. 
STNet and BLEEP encode the spot images using ResNet~\cite{16cvpr-resnet} while HistoGene uses ViT~\cite{vit} to encode the whole slide image.
Both STNet and HistoGene use multiple-layer perceptions (MLPs) to predict the gene expression values.
Differently, BLEEP replaces the MLPs with a retrieval database for the predictions.
Although those works can perform well on various datasets, they all learn a regressor for each gene, ignoring the potential relationship between genes.
Besides, those works can only make predictions for genes that appear in the training data and are unable to predict unseen genes.
Therefore, those limitations significantly hinder the model's generalization and flexibility.

To address the above limitations, this paper proposes GeneQuery, a simple yet effective question-answering-based framework for spatial gene expression prediction from histology whole slide images (WSI). 
Specifically, GeneQuery regards spot images of the histology images as contexts and gene metadata information as queries.
GeneQuery introduces the gene random variable to approximate the gene distribution for better model generalization.
This paper also presents two specific architecture implementations, i.e., spot-aware GeneQuery taking spot images as input sequences for capturing patterns across images, and gene-aware GeneQuery taking gene metadata as input sequences for capturing patterns between genes.
This paper evaluates GeneQuery with various comparisons on different datasets, including the challenging human liver tissue dataset captured via the 10x Visium platform and the human breast dataset HER2+ and HBD. 
Results show that GeneQuery can outperform state-of-the-art works and predict the gene expressions of unseen genes with comparable performance. 
Comprehensive analysis experiments are conducted to show the efficacy of the proposed GeneQuery.
Codes are available at \footnote{https://github.com/xy-always/GeneQuery}.

In summary, our contributions are as follows:
\begin{enumerate}
    \item This paper first defines the gene prediction problem as a question-answering task for better model generalization and flexibility.
    \item The proposed GeneQuery can capture different patterns between images and genes, images and images, and even genes and genes with two customized architectures, i.e., spot-aware GeneQuery and gene-aware GeneQuery.
    \item Experimental results demonstrate that the proposed GeneQuery can achieve state-of-the-art results on multiple ST datasets and achieve a competitive performance on unseen genes and in transfer learning scenarios.
\end{enumerate}

\section{Related Works}
\label{rel}

Recent advancements have leveraged histology images to predict gene expression profiles, showing promising outcomes for various downstream tasks. A pioneering study, HE2RNA~\cite{HERNA}, utilized these images to estimate gene expression across whole slide images, demonstrating the potential to integrate histological and gene expression analyses to enhance molecular mechanisms studies.
STNet~\cite{stnet}, a classical work in this field, employed a pre-trained ResNet to encode H\&E stained spot images, combined with a fully connected layer to predict gene expression values. Another notable project, HistoGene~\cite{histogene}, utilized a pre-trained VIT to encode H\&E spots, incorporating spatial information into each spot and using an MLP for prediction. Xie et al.~\cite{bleep} introduced a novel contrastive learning approach to develop a joint representation of H\&E images and gene expression profiles. Their model, BLEEP, predicts gene expression profiles by linearly combining the closest anchors from a reference database. However, STNet and HistoGene treat this task as a multi-output regression task, exploring different transformations to learn the relationship between image representations and gene expression, ignoring the relationship between different genes. BLEEP navigates the joint space of images and expressions of genes of interest. This method first needs to determine the number of genes during training so it can only predict predetermined genes.
In this work, we introduce GeneQuery, which reformulates this gene expression prediction problem by incorporating gene random variables and introduces rich gene metadata information, increasing the flexibility and robustness of the model.


\section{GeneQuery}
\label{met}

\begin{figure*}[t]
\centering
\includegraphics[width=0.8\linewidth]{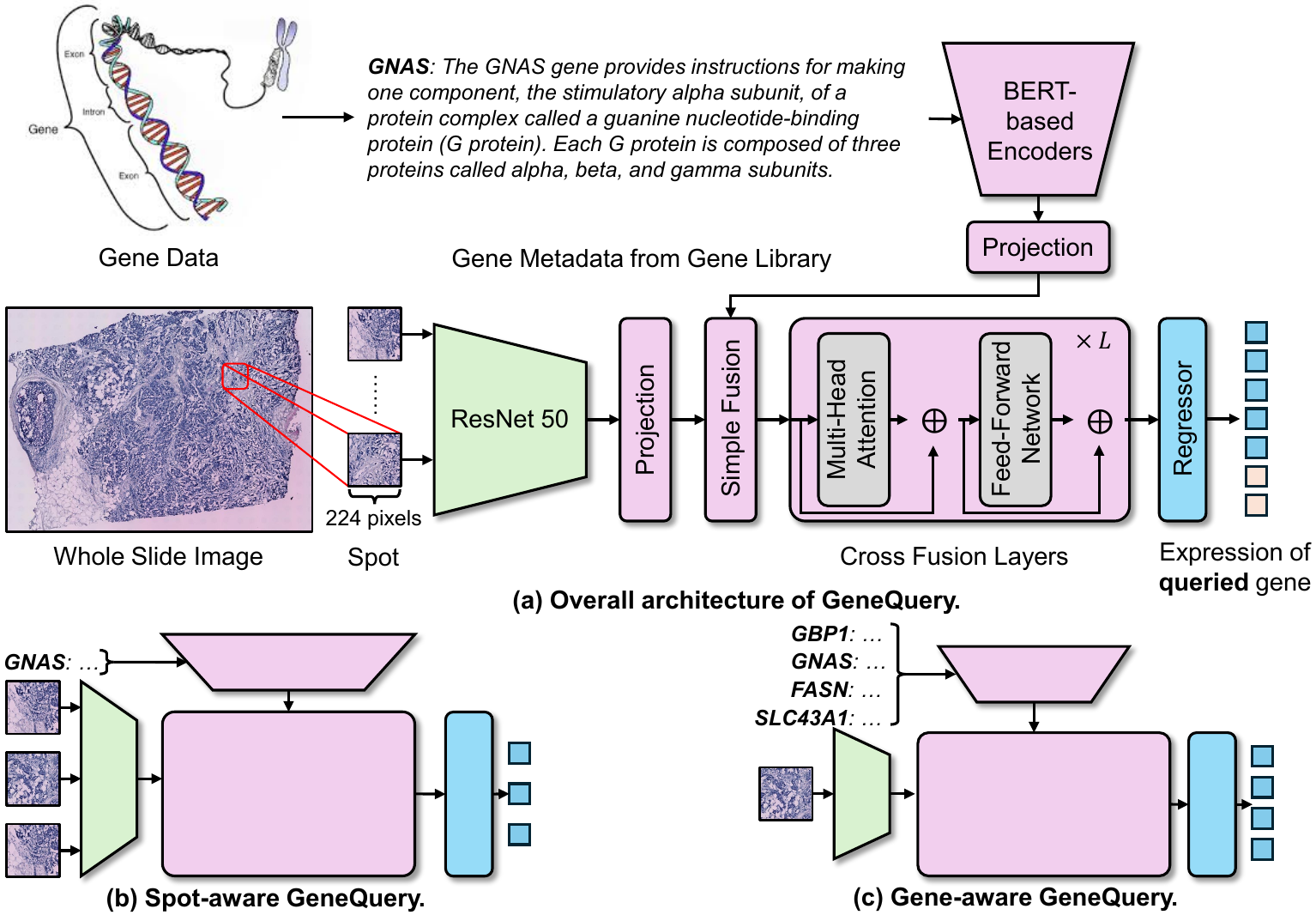}
\caption{The overall architecture of GeneQuery. (a) GeneQuery takes gene metadata as queries and spot images as contexts, then predicts the gene expression values. (b) Spot-aware GeneQuery takes the feature of spot images as the input sequence. (c) Gene-aware GeneQuery takes the feature of a list of genes as the input sequence.}
\label{fig:genequery}
\end{figure*}
In this section, this paper describes the proposed GeneQuery in detail.
Figure~\ref{fig:overview} (a) shows that existing frameworks~\cite{bleep, histogene, stnet} regard the gene predictions as a multi-regression problem and directly learn the pattern between images and genes. 
Differently, the proposed GeneQuery first re-formulates this multi-output regression problem as a more general question-answering problem, i.e., giving the whole slide image as context, querying any gene on it, and predicting the corresponding gene expression values.
Then, this paper presents the detailed techniques of the proposed GeneQuery framework in Figure~\ref{fig:overview} (b), including two specific implementations for capturing different data patterns, i.e., spot-aware GeneQuery and gene-aware GeneQuery.

\subsection{Problem Reformulation}
The formal definition of the gene expression prediction is that given a histology image that is partitioned into $n$ spots $X=\{x_1, \ldots, x_n\}, x_i\in\mathbb{R}$, it requires to design a novel framework to predict gene expression values of $k$ genes $g_1, \ldots, g_k$ for each spot $x_i$.
Let $y_{ij}$ represent the gene expression value of the gene $g_j$ on the spot $x_i$, $y_{ij}\in\mathbb{R}$.
Therefore, the goal is to estimate $k$ probability function $P(y_j\;|\;x_i)$ based on the training data.

For each gene $g_j$, existing works~\cite{stnet, histogene} leverage a model $\mathcal{M}$ to estimate the probability function $P(y_j\;|\;x_i)$ with parameters $\theta_j$~\footnote{Although there is only one model in their works, the weights for predicting different genes are not shared. Thus, parameters can be partitioned into multiple subsets $\{\theta_1, \ldots, \theta_k\}$, where $\theta_j$ is used for the regression problem of the gene $g_j$.} to make the predictions for the spot $x_i$, 
\begin{equation}
    P(y_j\;|\;x_i)\propto\mathcal{M}(x_i;\;\theta_j).
\end{equation}

\textbf{Limitations of Existing Works.} 
In this gene expression prediction problem, there are three kinds of implicit patterns between images and genes, images and images, and genes and genes.
Existing works mainly focus on the former two patterns, such as adopting a CNN-based neural network for the pattern between images and genes~\cite{stnet} or a transformer-based neural network for capturing the pattern between images and images~\cite{histogene}.
Such approaches are based on the assumption that all genes are independent, where each gene prediction problem is an independent regression problem and can be solved by estimating the corresponding probability function. 
However, those works ignore the relationships between genes and genes.
In biology, the expression of different genes often has a certain degree of synergy or interaction, especially in some diseases (such as cancer), where the expression patterns are interdependent~\cite{09NRG-gene-interaction}.
Besides, such methods would also be limited by the number of genes in the training data, while the probability function for new genes would be trained from scratch.

\textbf{Reformulated Gene Prediction.} 
To address the above limitations, this paper first models the gene as a random variable $g$, taking discrete values of $\{g_1, \ldots, g_m\}$, where $m$ represents the number of genes in the gene library.
Thus, the goal is reformulated to estimate the distribution of the gene expression values given the spot $x$ and the gene library, i.e., $P(y\;|\;x,\;g)$.
Then, we can use the Maximum Likelihood Estimation (MLE) to maximize the expectation of $P(y\;|\;x,\;g)$ on the training data, including the spot image $x$, the gene information $g$, and the gene expression values $y$,
\begin{equation}
\begin{aligned}
    \max \mathbb{E}_{x\sim X}\mathbb{E}_{y,g\sim G_x} P(y\;|\;x, g) \\
    = \min_{\theta=\{\phi, \rho\}}\sum^n_{i=1}\sum^k_{j=1}\mathcal{L}\left(y_{ij},f_{\phi}(x_i, g_j)\right),
\end{aligned}
\end{equation}
where $G_x=\{y_j, g_j\}^k_{j=1}$, $f(\cdot)$ is the model for predicting the gene expression values, $\theta$ represents all parameters in the whole framework.

Introducing the gene random variable can help the model capture the implicit patterns between different genes.
Unlike existing frameworks, the proposed GeneQuery directly learns the image distribution and gene distribution, enhancing the model's generalization ability and flexibility.
As shown in Figure~\ref{fig:genequery} (b) and (c), given different queries (genes) and contexts (images), GeneQuery can capture different patterns across images and genes.
When predicting gene expression values for a specific gene $g_{k}$, GeneQuery sets the gene random variable $g$ as $g_{k}$, i.e., $P(y\;|\;x,\;g=g_{k})$ (Figure~\ref{fig:genequery} (b));
While only one image $x_i$ is given and multiple genes' expression values are required to be predicted, GeneQuery sets the image input, i.e., $P(y\;|\;x=x_i,\;g)$ (Figure~\ref{fig:genequery} (c)).

\subsection{GeneQuery}
As shown in Figure~\ref{fig:genequery} (a), this section presents the overall architecture of GeneQuery.
GeneQuery takes the gene metadata as queries, the spot images as contexts, and predicts the gene expression values. 
Gene metadata can be any information related to genes, such as gene names or gene descriptions in the gene library, or even generated content by generative models.
GeneQuery basically consists of two encoder models $E_{img}, E_{gene}$ for encoding image features and gene features, two projection layers $M_{img}, M_{gene}$ for aligning the feature dimension, one fusion module $F$ for fusing image features and gene features, and one regressor $R$ for the final predictions.

To obtain the gene expression value, GeneQuery first encodes the spot of the histology image and the gene metadata separately,
\begin{equation}
    e_{img}=E_{img}(x_i),\quad e_{gene}=E_{gene}(g_j).
\end{equation}
The encoder models can be any pre-trained models, such as ResNet~\cite{16cvpr-resnet} for image encoders and clinical BERT~\cite{23nm-clinical-bert} for gene metadata encoders.
Then, GeneQuery leverages two projection layers to align the gene features and image features into the same dimensions since different encoders may produce different feature dimensions.
\begin{equation}
    h_{img} = M_{img}(e_{img}), \quad h_{gene} = M_{gene}(e_{gene}).
\end{equation}
In the fusion module, GeneQuery first uses a simple fusion layer $F_{joint}$ for naively fusing two kinds of features,
\begin{equation}
    h_{joint}=F_{joint}(h_{img}, h_{gene}).
\end{equation}
After obtaining the joint representation $h_{joint}$ for the images and the genes, GeneQuery uses $L$ transformer blocks $F^{l}_{trans}$ to cross-fuse the image information and gene information,
\begin{equation}
    h^l_{joint} = F^l_{trans}(h^{l-1}_{joint}),
\end{equation}
where $h^1_{joint}=h_{joint}$.
Finally, GeneQuery predicts the gene expression values via a regressor $R$,
\begin{equation}
    \hat{y}_{ij}=R(h^L_{joint}).
\end{equation}

In the training stage, GeneQuery uses Mean Squared Error (MSE) loss to compare each predicted gene expression $\hat{y}_{ij}$ with true gene expression $y_{ij}$ and optimizes to minimize the loss over the training data including all spots and all genes,
\begin{equation}
    \mathcal{L} = \frac{1}{nk}\sum_{x_i\in X}\sum^k_{j=1} (y_{ij} - \hat{y}_{ij})^2.
\end{equation}

GeneQuery aims to capture all three kinds of implicit patterns between images and genes,  images and images, and genes and genes.
However, jointly learning all patterns across images and genes is not easy.
Although GeneQuery can take all spot images and all genes as image input sequences and gene input sequences, it would introduce significant training overheads and make the whole framework hard to converge.
Therefore, GeneQuery has two specific implementations, 
\begin{itemize}
    \item \textbf{Spot-Aware GeneQuery.} Although HistoGene~\cite{histogene} has proposed to use the vision transformer to capture the relationships between images and images, it only predicts gene values using a linear layer whose dimension is the number of genes. The proposed spot-aware GeneQuery also takes all spot images of a whole slide image as input sequence, then adds the queried gene's features with each spot feature,
    \begin{equation}
        h^i_{joint} = h^i_{img} + h_{gene},
    \end{equation}
    where $i$ indicates the $i$-th spot image.
    \item \textbf{Gene-Aware GeneQuery.} To capture genes' relationships, gene-aware GeneQuery takes all genes as input sequences, then adds the feature of the given context (the spot image) on each gene feature,
    \begin{equation}
        h^j_{joint} = h_{img} + h^j_{gene},
    \end{equation}
    where $j$ indicates the $j$-th queried gene.
\end{itemize}

\section{Experiments}
\label{exp}

In this section, this paper first introduces the dataset and details its implementation.
Then, this paper presents the main results of predicted gene expressions.

\begin{table}[t]
  \caption{Statistics of datasets in terms of total number of whole-slide images, average number of spots per whole-slide image, number of genes in each dataset, and number of subtypes in each dataset.}
  \label{tab0}
  \centering
  \begin{tabular}{lccccc}
    \toprule
    Datasets & \# WSIs & \# Spots & \# Genes & \# Sub\\
    \midrule
    GSE  & 4 & 2317 & 3467 & 1 \\
    HER2+& 36 & 348 & 785 & 1 \\
    HBD & 68 & 450 & 723 & 4 \\
    \bottomrule
  \end{tabular}
\end{table}

\begin{table*}[t]
  \caption{Pearson correlation of all genes (ALL), top 50 most highly expressed genes (HEG), and top 50 most highly variable genes (HVG) compared to ground truth expressions on the held-out dataset. Results are reported with mean and variance. \textbf{The results in bold} are the best, and \underline{the underlined results} are the second-best. }
  \label{tab1}
  \centering
  \begin{tabular}{lllllll}
    \toprule
    \multicolumn{2}{c}{Datasets} & $\text{STNet}$~\cite{stnet} & $\text{HistoGene}$~\cite{histogene} & $\text{BLEEP}$~\cite{bleep} & GeneQuery\_gene & GeneQuery\_spot \\
    \midrule
    \multirow{3}{*}{GSE} 
    & HEG & 0.126$\pm$0.005 & 0.072$\pm$0.018 & 0.175$\pm$0.016 & \underline{0.243$\pm$0.028} & \textbf{0.256$\pm$0.039} \\
    & HVG & 0.091$\pm$0.007 & 0.071$\pm$0.011 & 0.173$\pm$0.011 & \underline{0.230$\pm$0.028} & \textbf{0.238$\pm$0.037} \\
    & ALL & 0.031$\pm$0.009 & 0.003$\pm$0.001 & 0.014$\pm$0.003 & \textbf{0.054$\pm$0.008} & \underline{0.051$\pm$0.015} \\
    \midrule
    \multirow{3}{*}{HER2+} 
    & HEG & 0.298$\pm$0.032 & 0.063$\pm$0.016 & 0.298$\pm$0.090 & \underline{0.315$\pm$0.030} & \textbf{0.317$\pm$0.074} \\
    & HVG & 0.302$\pm$0.026 & 0.064$\pm$0.012 & \textbf{0.322$\pm$0.100} & \underline{0.318$\pm$0.025} & 0.315$\pm$0.079 \\
    & ALL & 0.136$\pm$0.011 & 0.016$\pm$0.008 & 0.120$\pm$0.051 & \underline{0.159$\pm$0.027} & \textbf{0.174$\pm$0.049} \\
    \midrule
    \multirow{3}{*}{HBD} & HEG & 0.189$\pm$0.019 & 0.053$\pm$0.012  & \underline{0.191$\pm$0.037} & \textbf{0.200$\pm$0.009} & 0.149$\pm$0.029 \\
    & HVG & \underline{0.209$\pm$0.017} & 0.051$\pm$0.012 & 0.208$\pm$0.050 & \textbf{0.212$\pm$0.010} & 0.152$\pm$0.032\\
    & ALL & \textbf{0.073$\pm$0.011} & 0.015$\pm$0.004& \underline{0.068$\pm$0.014} & 0.061$\pm$0.007 & \underline{0.068$\pm$0.021} \\
    \bottomrule
  \end{tabular}
\end{table*}

\begin{table*}[t]
  \caption{PCC results of spot-aware GeneQuery on unseen genes. `\%' represents the ratio of seen genes used for training. `Seen' refers to the results of the seen genes; `Unseeen' refers to the results of the unseen genes; `All' refers to the results of all genes.}
  \label{tab2}
  \centering
  \begin{tabular}{lcccccccccc}
    \toprule
    & & \multicolumn{3}{c}{Seen} & \multicolumn{3}{c}{Unseen} & \multicolumn{3}{c}{All} \\
    Datasets & \% & HEG & HVG & ALL &  HEG & HVG & ALL & HEG & HVG & ALL \\
    \midrule
    \multirow{3}{*}{GSE}
    & 20\% & 0.117 & 0.117 & 0.029 & 0.137 & 0.120 & 0.024 & 0.140 & 0.128 & 0.025 \\
    & 40\% & 0.149 & 0.150 & 0.031 & 0.163 & 0.157 & 0.027 & 0.181 & 0.168 & 0.027 \\
    & 60\% & 0.157 & 0.154 & 0.031 & 0.178 & 0.169 & 0.035 & 0.202 & 0.187 & 0.033 \\
    \midrule
    \multirow{3}{*}{HER2+} 
    & 20\% & 0.156 & 0.152 & 0.109 & 0.109 & 0.092 & 0.051 & 0.162 & 0.151 & 0.063 \\
    & 40\% & 0.261 & 0.249 & 0.130 & 0.131 & 0.131 & 0.057 & 0.255 & 0.253 & 0.098 \\
    & 60\% & 0.296 & 0.309 & 0.163 & 0.121 & 0.101 & 0.049 & 0.255 & 0.266 & 0.099 \\
    \bottomrule
  \end{tabular}
\end{table*}
\subsection{Experimental Setup}
\label{sec:dataset}
\noindent\textbf{Datasets.} This paper uses spatial transcriptomics data for training and testing. The statistics of datasets we used are listed in Table ~\ref{tab0}. 
The spatial transcriptomics data includes histology images split into spatially barcoded spots and the corresponding spatial gene expression data. 
Specifically, this paper uses three representative datasets: GSE240429 (GSE), HER2+, and HBD. 
The GSE240429~\footnote{https://www.ncbi.nlm.nih.gov/geo/query/acc.cgi?acc=GSE240429} includes human liver tissue from neurologically deceased donor livers suitable for transplantation that were OCT embedded, frozen, sliced with a cryostat and imaged using the 10x Genomics Visium platform \footnote{https://www.10xgenomics.com/products/spatial-gene-expression}. 
The HER2+ dataset \footnote{https://github.com/almaan/her2st} is human HER2-positive breast tumor ST data and is collected from 8 HER2-positive breast cancer patients. The histology images of HER2+ have lower spatial resolution than Visium.
The HBD dataset \footnote{https://data.mendeley.com/datasets/29ntw7sh4r/5} is breast cancer patients with luminal a, luminal b, triple-negative, and HER2-positive subtypes. 
For both datasets, GeneQuery takes a $224\times 224$-pixel patch of the image centered on each spot as the input image patch. 
For the GSE240429 dataset, following BLEEP~\cite{bleep}, this paper selects 1,000 highly variable genes in each histology image, with 3467 union genes to predict. 
For the HER2+ dataset, following HistoGene \cite{histogene}, this paper selects the top 1,000 highly variable genes in each tissue section and removes genes expressed in less than 1,000 spots across all tissue sections. 
As a result, we use a total of 785 genes in HER2+ to predict.
We applied the same genes as in HER2+ to the HBD dataset, removed any missing genes, and retained 723 genes.
Each spot is normalized by log normalization and min-max normalization.

\noindent\textbf{Metrics and Comparisons.} To evaluate the predicted gene expression profile, we use the Pearson Correlation Coefficient (PCC) to estimate the correlation between the predicted gene values and observed gene values. 
We evaluate the average PCC of the top 50 highly expressed genes (HEG) and the top 50 highly variable genes (HVG) following the setting of BLEEP~\cite{bleep}. 
Also, we compute the average PCC of all trained genes (ALL) following the setting of STNet~\cite{stnet} and HistoGene~\cite{histogene}, which demonstrates the model's ability to predict gene expression overall. 
If no additional clues are provided in this paper, HEG, HVG, and ALL represent these evaluation metrics.

\subsection{Implementation Details}
\label{sec:impl}
We used the ResNet50~\cite{16cvpr-resnet} as the image encoder and the clinical BERT~\cite{23nm-clinical-bert} as the gene encoder. 
For gene-aware GeneQuery, the max length of the transformer block for GSE240429, HER2+, and HBD datasets are 3467, 785, and 723, respectively. 
For spot-aware GeneQuery, the max length of the transformer block for GSE240429, HER2+, and HBD datasets are 2400, 600, and 600, respectively.
The transformer layer $L$ is set to 2.
The fusion dimension for gene and image representation is 256, the batch size is 100. 
We run 100 epochs and save the last checkpoint for testing. 
For GSE240429, because there is a total of four histology WSI, we did 4-fold cross-validation, and each fold left 1 histology image for testing and the remaining for training. 
For the HER2+ and HBD datasets, we did 5-fold cross-validation, using 20\% histology images for testing and the remaining for training. 
The evaluation set is randomly selected 10\% from the training set. 
For HistoGene and BLEEP, we use the default hyperparameters reported in their papers. 
For STNet, we implemented it on our own, and the hyperparameters are the same as reported in their paper. 
All experiments are conducted on a 40G A100 GPU and 32G V100 GPU.

\subsection{Main Results}
Table~\ref{tab1} presents the quantitative results and reports the mean and standard deviation of PCC results on the HEG, HVG, and ALL settings. 
As expected, gene-aware or spot-aware GeneQuery can achieve the best performance across all datasets and settings, except for competitive results on the HVG setting of HER2+ and the ALL setting of HBD. 
Specifically, gene-aware and spot-aware GeneQuery surpass the STNet, HistoGene, and BLEEP by 9.3\%, 12.7\%, 5.5\%, 9.9\%, 13.3\%, and 6.1\% on average over all settings on the GSE dataset, respectively. 
For the HER2+ dataset, gene-aware and spot-aware GeneQuery achieve higher 1.9\%, 21.6\%, 1.7\%, 2.3\%, 22.1\%, and 2.2\% on average compared to three baseline methods.
For the HBD dataset, gene-aware and spot-aware GeneQuery outperform HistoGene by 11.8\% and 8.3\% on average, respectively, but achieve competitive results compared to STNet and BLEEP.
Overall, the proposed GeneQuery can achieve a more general and stable performance compared to existing state-of-the-art works.
This may be due to the introduction of the gene random variable, which helps the GeneQuery framework capture the gene distribution across different datasets.

Comparing gene-aware and spot-aware GeneQuery, it is hard to conclude that one can consistently outperform the other.
Specifically, on the GSE dataset, spot-aware GeneQuery performs better than gene-aware GeneQuery over highly expressed and variable genes but competitively on the results of all genes.
On the HER2+ dataset, spot-aware GeneQuery achieves comparable results to gene-aware GeneQuery on HEG and HVG but performs better over all genes.
Therefore, spot-aware GeneQuery outperforms gene-aware GeneQuery on the first two datasets.
The reason might be that there is only one subtype of the WSI in those datasets, exhibiting relatively low heterogeneity of data, which makes it easier to capture stable features for image encoders compared with more subtypes of histology images.
However, when the number of subtypes increases (4 subtypes in the HBD dataset), spot-aware GeneQuery performs poorly, as different WSIs with different subtypes may show different patterns in tissue architecture and cell morphology.

Besides, the results of different methods on the HER2+ dataset are much better than those on the GSE dataset.
This is because there are fewer spots in each WSI in the HER2+ dataset and also fewer genes for predicting in the HER2+ dataset, which makes the models converge easily.
However, when the number of spots and genes increases, the proposed GeneQuery can still maintain an acceptable performance.
This benefits from the generalization ability of GeneQuery.

The above results demonstrate the efficacy of the proposed GeneQuery on the gene expression profiling prediction tasks. 
However, the absolute PCC values still remain relatively low, indicating the difficulty of the gene expression prediction task from histology images. 
It shows that some genes may have weak relations with histology images, as caused by inadequate detection of specific genes by the gene detection platform such as the Visium platform, leading to less predictable expression patterns and experimental artifacts and introducing non-biological variation in the data unrelated to the image.

\subsection{Results on Unseen Genes}
Different from existing works, predicting unseen genes is a unique feature of the proposed GeneQuery framework. 
Table~\ref{tab2} shows the results of spot-aware GeneQuery on different ratios of unseen genes on two representative datasets and all three settings. 
We also report results of seen genes, unseen genes, and all genes.

From the experimental results in Table~\ref{tab2}, we can draw the conclusion that the proposed GeneQuery can still achieve acceptable performance on unseen genes, even when it is trained on very few genes.
Specifically, on the GSE dataset, as expected, when the number of seen genes increases, the spot-aware GeneQuery's performance on unseen genes improves.
Similar patterns also appear in the HER2+ dataset. 
This is because more genes expose models to more biological features and patterns, and learn richer and more comprehensive expressions.

Surprisingly, the proposed GeneQuery can achieve a better Pearson correlation on unseen genes than seen genes on the GSE dataset.
However, the unseen-gene-prediction results on the HER2+ dataset are the opposite, which are much lower than that on seen genes.
The reason may lie in the fact that the number of genes in the GSE dataset is much larger than the HER2+ dataset's number of genes, although both datasets all have one subtype of images.


\begin{table}
  \caption{Transfer learning results of spot-aware GeneQuery. Results are reported with mean and variance. The results of the first five rows are evaluated across the different tissues, while the results of the last three rows are evaluated within the same tissue.}
  \label{tab3}
  \centering
  \begin{tabular}{lllll}
    \toprule
    Train & Test & HEG & HVG & ALL\\
    \midrule
    GSE&HER2+ & $\text{0.108}_\text{0.072}$ & $\text{0.117}_\text{0.071}$& $\text{0.090}_\text{0.041}$  \\
    GSE&HBD& $\text{0.096}_\text{0.063}$ & $\text{0.095}_\text{0.066}$ & $\text{0.048}_\text{0.029}$  \\
    HER2+&GSE & $\text{0.039}_\text{0.015}$ & $\text{0.039}_\text{0.016}$ & $\text{0.009}_\text{0.005}$  \\
    HBD&GSE & $\text{0.014}_\text{0.009}$ & $\text{0.012}_\text{0.009}$ & $\text{0.003}_\text{0.001}$ \\
    \multicolumn{2}{c}{Average} & 0.064 & 0.066 & 0.038 \\
    \midrule
    HER2+&HBD &$\text{0.131}_\text{0.047 }$& $\text{0.133}_\text{0.047}$ &$\text{0.066}_\text{0.018}$  \\
    HBD&HER2+  & $\text{0.112}_\text{0.038}$ &$\text{0.107}_\text{0.036}$  & $\text{0.071}_\text{0.018}$  \\
    \multicolumn{2}{c}{Average} & 0.122 & 0.120 & 0.069 \\
    \bottomrule
  \end{tabular}
\end{table}

\subsection{Transfer Learning Results}
To show the generalization ability of the proposed GeneQuery framework, we conducted a transfer learning experiment on different datasets in terms of the same tissue and different tissues in Table~\ref{tab3}.
We also chose the spot-aware GeneQuery for its good performance on most datasets.
Experimental results show that the proposed GeneQuery exhibits a better transfer ability within the same tissues than that across different tissues.
The spot-aware GeneQuery especially achieves the best transfer performance between HER2+ and HBD datasets.
Although the transferred results of spot-aware GeneQuery are much lower than those of STNet, BLEEP, gene-aware GeneQuery, and its own, it can still achieve a better performance than that of HistoGene.
This also demonstrates the efficacy of the question-answering framework in the spot-aware GeneQuery compared to traditional solutions of multi-output regression.
Besides, for the transfer learning across different tissues, the models trained on the GSE datasets perform significantly better than those on the HER2+ or HBD datasets.
This is because there are more genes in the GSE dataset, and GeneQuery's generalization ability guarantees an acceptable performance.
It is also worth noting that the correlation of the predicted genes is significantly positively correlated, and no gene is significantly negatively correlated, which shows the potential of the proposed GeneQuery framework.

\subsection{Enhancing GeneQuery with GPT-4}

\begin{table}[t]
  \caption{Results of the proposed GeneQuery with the GPT-4 enhanced gene metadata. Results are evaluated on one fold. \textbf{The results in bold} are the best, and \underline{the underlined results} are the second-best.}
  \label{tab4}
  \centering
  \begin{tabular}{lllll}
    \toprule
    $D$ & Model& HEG & HVG & ALL \\
    \midrule
    \multirow{4}{*}{\rotatebox{90}{GSE}} 
    & GeneQuery\_gene & 0.271 & 0.258 & 0.062 \\
    & \quad w/ GPT-4 & \underline{0.282} & \underline{0.260} & \textbf{0.077} \\
    & GeneQuery\_spot & 0.272 & 0.258 & 0.046 \\
    & \quad w/ GPT-4 & \textbf{0.312} & \textbf{0.282} & \underline{0.068} \\
    \midrule
    \multirow{4}{*}{\rotatebox{90}{HER2+}} 
    & GeneQuery\_gene & 0.299 & 0.299 & 0.104 \\
    & \quad w/ GPT-4 & 0.333 & 0.327 & \textbf{0.199} \\
    & GeneQuery\_spot & \textbf{0.342} & \textbf{0.347} & \underline{0.183} \\
    & \quad w/ GPT-4 & \underline{0.336} & \underline{0.341} & 0.164\\
    \midrule
    \multirow{4}{*}{\rotatebox{90}{HBD}} 
    & GeneQuery\_gene & \underline{0.209} & \underline{0.222} & \underline{0.068} \\
    & \quad w/ GPT-4 & \textbf{0.257} & \textbf{0.278} & \textbf{0.100} \\
    & GeneQuery\_spot & 0.120 & 0.124 & 0.046 \\
    & \quad w/ GPT-4 & 0.161 & 0.168 & 0.072 \\
    \bottomrule
  \end{tabular}
\end{table}

\begin{figure*}[t]
\centering
\includegraphics[width=1.0\linewidth]{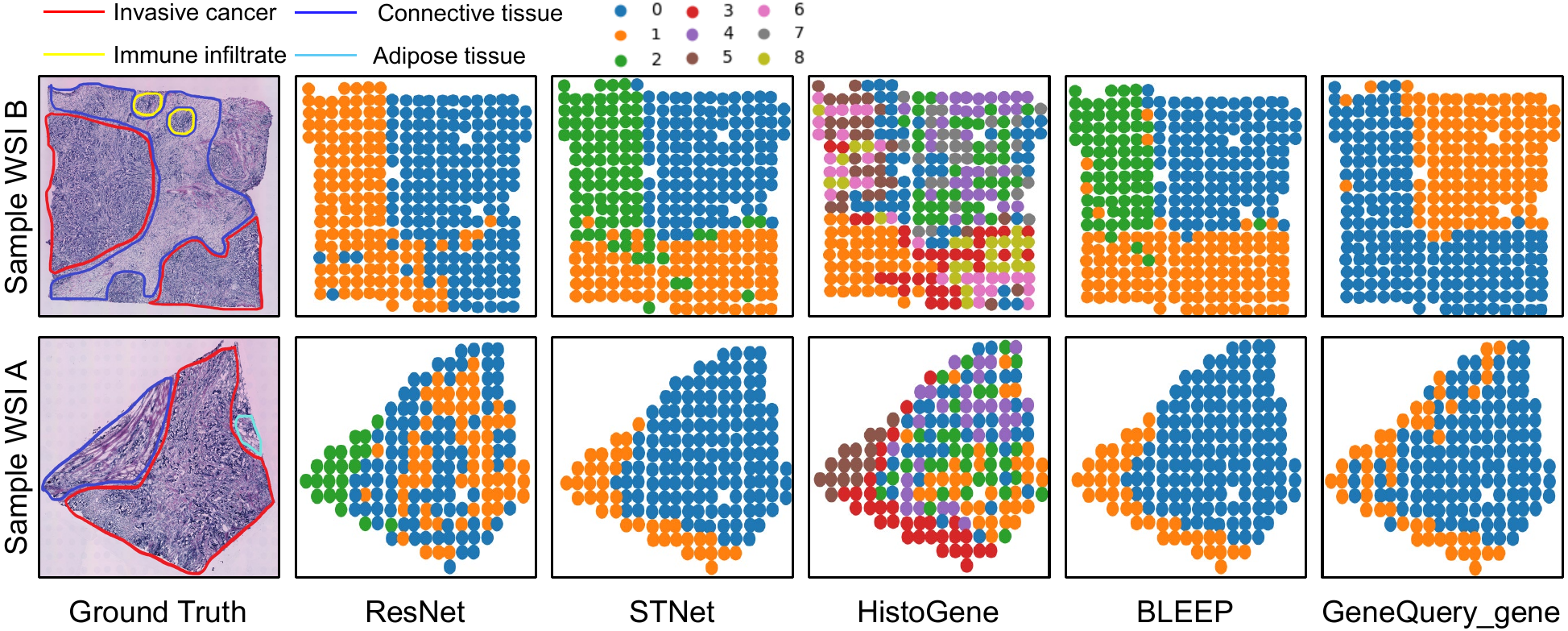}
\caption{Visualization of spot latent representation on HER2+ dataset. Visualization results of different methods are shown in spots, where each spot is colored according to the unsupervised clustering algorithms.}
\label{fig:pear}
\end{figure*}

Since some gene metadata is too simple, we conducted an experiment using the state-of-the-art large language model GPT-4 to generate comprehensive gene descriptions to enhance the GeneQuery.
The prompt template used for GPT-4 is ``The brief definition of gene \texttt{GENE\_NAME} is '', where \texttt{GENE\_NAME} is the gene name in the gene library. 
For each dataset, we conducted validation experiments on one fold. 
Experimental results in Table~\ref{tab4} show that with the rich gene metadata generated by GPT-4, gene-aware GeneQuery and spot-aware GeneQuery can be improved by 3.6\% and 1.8\% on average over all datasets and settings, respectively.
Specifically, GPT-4 helps gene-aware and spot-aware GeneQuery enhance the average performances on GSE, HER2+, and HBD by 0.9\%, 5.2\%, 4.5\%, 2.9\%, and 3.7\%, respectively, except for the competitive results of spot-aware GeneQuery on the HER2+ dataset.
The above results suggest that generated gene information may cover rich biology knowledge or attributes, enabling models to better understand the biological semantics of queries and make more accurate predictions from the histology images.

\subsection{Latent Image Space of GeneQuery}
This section visualizes the latent spots histology image representation of the gene expression models as we can inspect the quality and biological relevance of the features the gene expression models have learned. 
We use Uniform Manifold Approximation and Projection (UMAP)~\cite{umap} to perform the feature dimension reduction and then utilize Leiden~\cite{leiden} to cluster spot representations. 
Additionally, segmentation can be performed based on the morphological features of the images, we also utilized a pre-trained ResNet50 (See ResNet in Figure~\ref{fig:pear}) model to represent the images, followed by the same dimensionality reduction and clustering techniques.

As shown in Figure~\ref{fig:pear}, most gene expression models, including STNet, BLEEP, and GeneQuery, can effectively segment histology images. It is noteworthy that GeneQuery demonstrates promising potential for segmenting histology images. 
Specifically, the gold standard annotations for Sample WSI A primarily identify three regions: invasive cancer, connective tissue, and adipose tissue. 
ResNet, STNet, and BLEEP are all able to distinguish invasive cancer and connective tissue.
However, ResNet further segmented the invasive cancer region into sub-regions.
STNet and BLEEP miss the top areas of connective tissue.
In contrast, GeneQuery is able to identify a greater extent of the connective tissue region. 

For Sample WSI B, the gold standard annotations mark three regions: immune infiltrate, invasive cancer, and connective tissue. 
ResNet is unable to distinguish between invasive cancer and connective tissue on the right side of the sample. 
Although STNet and BLEEP could differentiate these regions, they label the same invasive cancer area as different categories. 
GeneQuery outperforms all other models, successfully distinguishing between invasive cancer and connective tissue while also recognizing all invasive cancer regions. 
The segmentation results indicate that with the exception of HistoGene, which was particularly challenging to segment, all other models successfully performed segmentation with clear outlines. 
These findings underscore the robustness of the models, especially GeneQuery, in effectively segmenting histological data.
This indicates that GeneQuery may be able to integrate rich gene metadata to help the model better capture the heterogeneity of tissue images. 
This capability may offer significant support for future cancer diagnostics.

\section{Conlusion}
\label{con}

This paper proposes a flexible and general question-answering-based framework named GeneQuery, which predicts gene expression profiles from histology images.
The proposed GeneQuery reformulates the gene expression prediction problem by introducing the gene random variable.
The proposed GeneQuery takes rich gene metadata as queries and obtains the answers (gene expression values) from corresponding histology whole slide images.
The proposed two architecture implementations, i.e., spot-aware GeneQuery and gene-aware GeneQuery, demonstrate impressive performance on not only known genes but also previously unseen genes.
Experiments also exhibit GeneQuery's transfer learning capability.
With such a QA-based framework, GeneQuery can integrate various types of gene meta-information and flexibly predict any gene expression values from the whole slide images. 
This paper demonstrates GeneQuery's robust ability to handle multi-modal data, offering a novel technique for genomic fields to drive advances in gene expression research.
\section{Acknowledgements}
\label{ack}

We acknowledge Tianle Zhong and Qiang Su for their computing resource support.
\bibliographystyle{unsrt}  
\bibliography{references}

\end{document}